\documentclass[runningheads]{llncs}
\usepackage{graphicx}
\usepackage{amsmath,amssymb} 

\usepackage{algorithm}
\usepackage[noend]{algpseudocode}
\algnewcommand\algorithmicinput{\textbf{Input:}}
\algnewcommand\Input{\item[\algorithmicinput]}

\algnewcommand\algorithmicoutput{\textbf{Output:}}
\algnewcommand\Output{\item[\algorithmicoutput]}

\usepackage{color}
\usepackage{subcaption}
\usepackage{wrapfig,lipsum,booktabs}

\usepackage{capt-of}

\usepackage{multirow}

\usepackage{float}
\usepackage{dblfloatfix}
\usepackage{array}
\newcolumntype{P}[1]{>{\centering\arraybackslash}p{#1}}

\usepackage{color,soul}

\usepackage{lineno,hyperref}
\usepackage{amsmath}
\newcommand{\argmin}{\arg\!\min}
\modulolinenumbers[5]
\allowdisplaybreaks

\begin{document}

\title{Direct Visual-Inertial Odometry with Semi-Dense Mapping} 

\titlerunning{Direct Visual-Inertial Odometry}

\author{Wenju Xu\and
		Dongkyu Choi\and
		Guanghui Wang}
%
\authorrunning{Wenju \it{et al.}}
%

\institute{School of Engineering, University of Kansas, Lawrence, KS, USA 66045}

\maketitle
\begin{abstract}
The paper presents a direct visual-inertial odometry system. In particular, a tightly coupled nonlinear optimization based method is proposed by integrating the recent advances in direct dense tracking and Inertial Measurement Unit (IMU) pre-integration, and a factor graph optimization is adapted to estimate the pose of the camera and rebuild a semi-dense map. Two sliding windows are maintained in the proposed approach. The first one, based on Direct Sparse Odometry (DSO), is to estimate the depths of candidate points for mapping and dense visual tracking. In the second one, measurements from the IMU pre-integration and dense visual tracking are fused probabilistically using a tightly-coupled, optimization-based sensor fusion framework. As a result, the IMU pre-integration provides additional constraints to suppress the scale drift induced by the visual odometry. Evaluations on real-world benchmark datasets show that the proposed method achieves competitive results in indoor scenes.
\end{abstract}




\section{INTRODUCTION}

In the research area of robotics, camera motion estimation and 3D reconstruction take fundamental places both for navigation and perception, such as unmanned aerial vehicle (UAV) navigation \cite{OKVIS} and indoor reconstruction \cite{engel2014lsd,engel2016direct}. Among these applications, an up-to-scale camera motion tracking and 3D structure of the environment are required at the end. Most existing methods formulate this problem as simultaneously localization and mapping (SLAM), which characterized on the sensors it used. Recent efforts include visual SLAM and visual inertial navigation system (VINS).

Visual odometry \cite{newcombe2011} estimates the depth of features, based on which, track the pose of the camera. In contrast, direct visual odometry without the feature processing pipeline eliminates some of the issues in feature
based methods through pose tracking directly based on pixels. However, even these methods are subject to 
failure during aggressive motions as images can be severely 
blurred and extremely sensitive to noise, changing illumination, and fast motion. Consequently, aggressive motion of the UAV \cite{ling2015,ling2016} with significant large angular velocities and linear accelerations, and lighting condition variation make the state estimation subject to scale drift in long term. 

While IMU generate noisy but outlier-free measurements, making them great for tracking of short-term fast motions. 
However, low-cost MEMS IMUs suffer significant drift in 
the long term. IMU measurements can improve visual odometry to remedy this issue by providing a short term measurement, making them an ideal combination to better estimate the pose of the camera and generate
information on the surroundings. We believe that integrating the complementary natures of the dense visual tracking and IMU measurements 
opens up the possibility of reliable tracking of aggressive 
motions.  

In this paper, we propose a tightly coupled function that integrates visual and inertial terms in a fully probabilistic manner. We adopt the concept of keyframes
due to its successful application in classical vision-only approaches: it is implemented using partial linearization and marginalization. The keyframe paradigm also
accounts for drift-free estimation when slow or no
motion is present. Instead of using an optimization
window of time-successive poses, the selected keyframes may
be spaced arbitrarily far in time, keeping visual constraints,
while still incorporating an IMU term. We provide a strictly probabilistic derivation of the IMU error terms and the respective information
matrix, relating successive image frames without explicitly
introducing the states at the IMU rate. 

The key contribution of this work is a robust and fully 
integrated system for direct visual inertial odometry. The novelties of the proposed system include:
1) The combination of the direct photometric information and the edge features, which are points with sufficiently high image gradient magnitude. We work with pixels' intensities, which provides good invariance to changes in viewpoint and illumination. This makes the system more robust and reliable than other methods dealing with detected features. 
2) IMU pre-integration. The IMU pre-integration provides scale information by integrating IMU measurements. Thanks to the
use of a factor graph, tracking and mapping are focused in a local covisible area, independent of global map
size.
3) tightly coupled optimization. The measurements from the IMU pre-integration and dense visual tracking are fused probabilistically using a tightly-coupled, optimization-based sensor fusion framework. In this paper, the dense visual tracking results provide the visual constraints between current frame and reference frame. While the IMU pre-integration provides constraints between two consecutive frames.
\begin{figure}[t!]
	\begin{center}
		
		\includegraphics[width=1\textwidth]{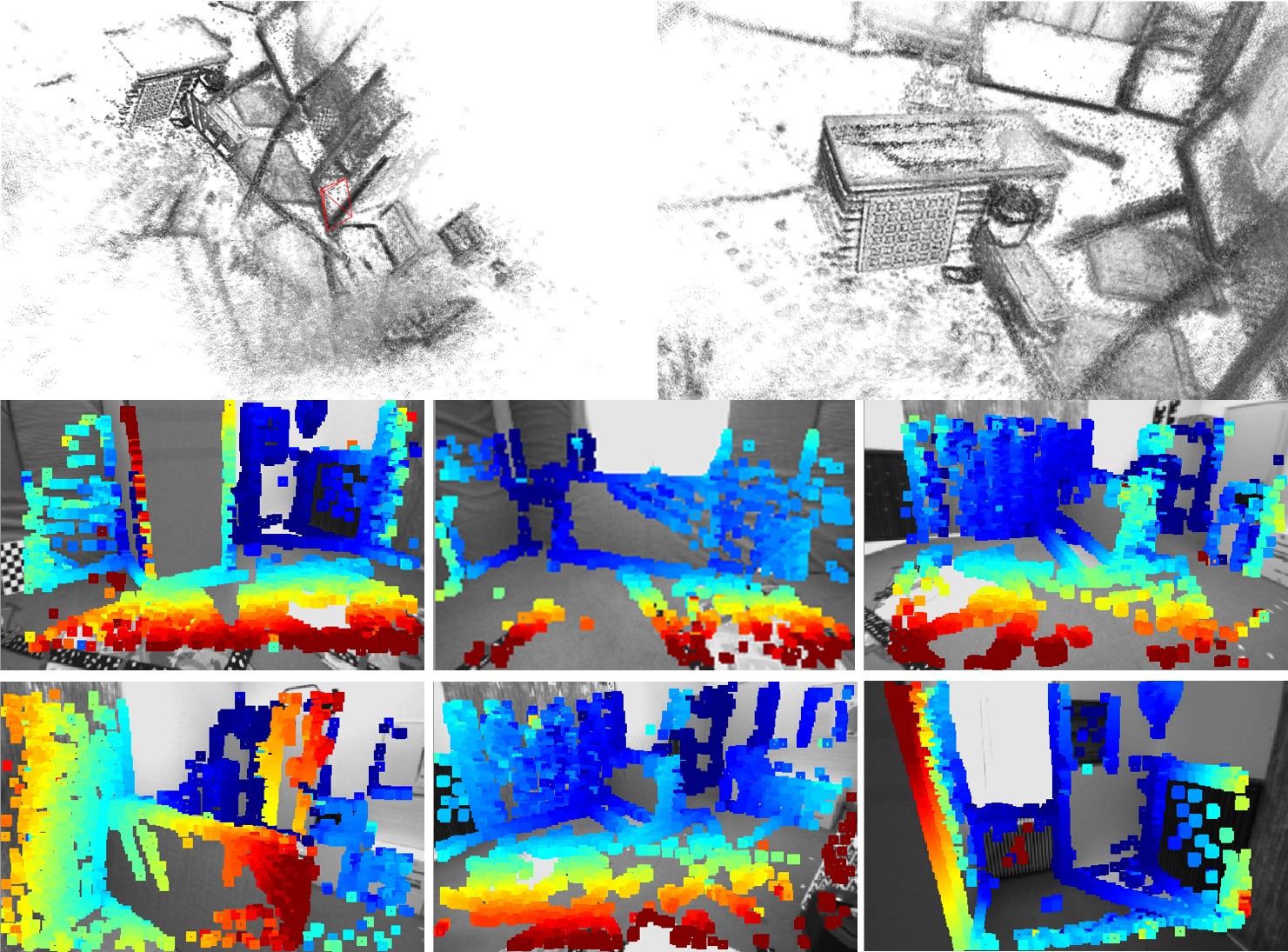}
	\end{center}
	\caption{3D reconstruction and depth estimation on EuRoC dataset. The first row shows the 3D reconstruction on the V1\_easy sequence and the bottom rows show the depth maps for frame tracking.}
	\label{f1}
\end{figure}
In the remainder of the paper, we begin with a review of the related work. In Sect. {\ref{overview}} An overview of the system is described and the details of IMU and visual measurement are are then presented. Dense mapping is introduced in Sect. {\ref{densemapping}}. Sect. {\ref{implement}} shows implement details. The results of experiment are discussed in Sect. {\ref{experiment}}. Finally, Sect. {\ref{conclusion}} draws conclusions.

\section{Related work}
Simultaneously localization and mapping has a long history in a monocular scenario, prominent examples include DTAM {\cite{newcombe2011}}, SVO {\cite{forster2014svo}} and LSD-SLAM {\cite{engel2014}}, which works on sparse features and estimate the camera motion through a prediction and correction fashion. Direct methods operating directly on image intensities have become popular since it is robust and computational efficient without feature detection. It is demonstrated that direct methods are suitable for dense mapping than feature based method, and when enhanced by edge alignment {\cite{kuse2016robust}} {\cite{Wang2016BMVC}}, it can deal with changing illumination and fast motion. More recently, direct sparse odometry (DSO) {\cite{engel2016direct}} presented an impressive semi-dense 3D reconstruction of the environment through a sliding window optimization method. This direct method minimizes the photometric error of points with sufficiently high image gradient magnitude (edges) using a non-linear optimization framework.

More recently, Deep neural network demonstrate advanced performance \cite{liu2019approximate}. People pay great attention to the 3D SLAM, like the semantic SLAM \cite{zhao2017fully} and the deep learning enhanced SLAM \cite{tateno2017cnn}. Semantic SLam \cite{zhao2017fully} focuses on simultaneous 3D reconstruction and material recognition and segmentation, which ends up with a real-time end-to-end system. Deep learning is promising in computer vision \cite{zhou2014smart,XU2019195,XU2019570,xu2019}. \cite{tateno2017cnn} takes the advantage of depth prediction from Convolutional Neural Networks (CNNs) to enhance the performance of monocular slam. DSO {\cite{engel2016direct}} takes one monocular camera for 3D reconstruction within inverse depth estimation framework. It is based on photometric error optimization of windowed sparse bundle adjustment.
	
All of these methods try to explore the scene within the reconstructed 3D map, while one problem they suffer is the scale drift.
There are two main approaches towards
solving the scale drift problem with extra measurements: the batch
nonlinear optimization
methods and the recursive
filtering methods. The former one jointly minimizes the error originated from the integrated IMU measurements and the (reprojection) errors from the visual terms \cite{OKVIS}; while the
recursive algorithms usually utilize the IMU measurements
for state propagation while the updates come from the visual
observations \cite{sibley2010,li2013}. At the back-end, fusion methods can largely be
divided into two classes: loosely-coupled fusion \cite{weiss2012real,scaramuzza2014vision} and tightly-coupled fusion \cite{concha2016,usenkodirect,ling2016,OKVIS}. In loosely-coupled fusion,
visual measurements are first processed independently to
obtain high-level pose information and then fused with
the inertial  measurements, usually using a filtering framework
\cite{li2013,li2012improving}. The two sub-problems are solved separately in a loosely-coupled fusion,
resulting in a lower computational cost, however, the results are
suboptimal. In tightly-coupled fusion, both the visual and the inertial measurements are fused and optimized in a
single framework. It considers the coupling between the two types of measurement and allows the adoption of a graph
optimization-based framework with iterative re-linearization
to achieve better performance. Tightly-coupled methods usually come with a higher computational cost.

\section{Problem Fromulation}
\subsection{Notation}
We use the following notation in the paper: 
\begin{itemize}
\item $\hat{m}^k_{k+1}$: the IMU measurements between the
$k$-th and $(k+1)$-th images;
\item  $\hat{m}^{ref(c)}_c$: the dense tracking
result for the current image $c$ with respect to the corresponding reference
image $ref(c)$ (a key frame);
\item  $r_{IMU}(\hat{m}^k_{k+1},\pi^k_{k+1})$: the
residual between the IMU integration and state
$\pi^k_{k+1}$;
\item $r_I(\hat{m}^{ref(c)}_c,\pi^{ref(c)}_c)$: the dense
tracking residual between $\hat{m}^{ref(c)}_c$ and state
$\pi^{ref(c)}_c$;
\item  $\Sigma_{IMU}$: the associated covariances of
the IMU measurement;
\item $\Sigma_I$: the associated covariances of the image alignment;
\item  $\pi^G_k$: the $k$-th state in the global coordinate system;
\item $p^G_k$: the $k$-th position states in the global coordinate system;  
\item $v^G_k$: the $k$-th velocity states in the global coordinate system; 
\item ${\theta}^G_k$: the $k$-th angular states in the global coordinate system; 
\item $b_a$: the $k$-th acceleration bias states in the global coordinate system;  
\item $b_g$: the $k$-th angular acceleration bias states in the global coordinate system;
\item  ${\theta}^G_k = log( {R}^G_k)^{\vee}$.
\end{itemize}

Our system maintains several states during the processing. A state
includes the position, velocity, orientation, accelerometer bias, and
gyroscope bias. The full state space is defined as:
\begin{align*}
\pi_k^G&=\begin{bmatrix}p^G_k & v^G_k & {\theta}^G_k & b_a & b_g\end{bmatrix}\\
\pi&=\begin{bmatrix}\pi^G_0 & \cdots & \pi^G_k & \cdots & \pi^G_n \end{bmatrix}
\end{align*}

\section{Visual-Inertial Odometry}\label{overview}
\begin{figure}[h]
	\begin{center}
		\includegraphics[width=0.9\textwidth]{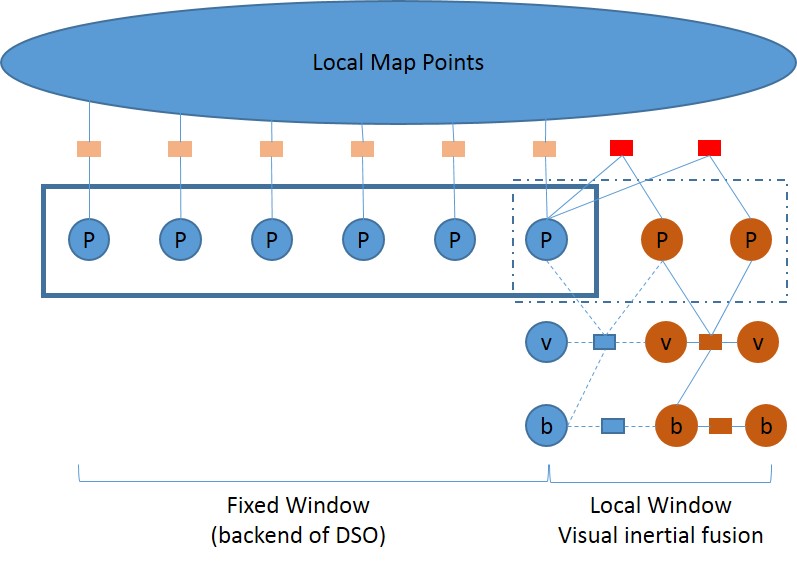}
	\end{center}
	\caption{Overview of the system. After the fixed window used by DSO, we maintain a local window that contains one keyframe and two latest regular frames. Based on the point cloud of the DSO, a visual inertial fusion system are proposed with the IMU pre-integration constraint and the frame tracking constraint. P stands for pose states. v stands for the velocity states. b stands for the bias states. Square represents the states and rectangle represents constraints between states.}
	\label{f2}
\end{figure}

In  this  section,  we  present  our  approach  of  incorporating  inertial  measurements  into  batch  visual  SLAM.  In  visual
odometry and SLAM, a nonlinear optimization is formulated to find the camera poses and landmark positions by minimizing
the reprojection error of landmarks observed in camera frames. Unlike in work {\cite{ling2016}}, which use stereo camera for depth estimation, we adopt the DSO to build a semi-dense map and take the depth estimation for camera motion tracking. Figure {\ref{f2}} shows the factor graph, in which the visual inertial fusion takes the points in local map as landmarks which provides one kind of geometric constraints to related observation states. The local map points are a collection of points with depth information in latest keyframes.

We seek to formulate the visual-inertial localization and mapping problem as one joint optimization of a cost function
$J(\pi)$ containing both the (weighted) reprojection errors
$r_I$ and the (weighted) IMU error $r_{IMU}$.
\begin{align}
\begin{split}
J(\pi):= &\sum_{k\in S_{IMU}} \Arrowvert
r_{IMU}\left(\hat{m}^k_{k+1},\pi^k_{k+1}\right)\Arrowvert^2_{\Sigma_{IMU}}\\
&+ \sum_{c\in C_I} \Arrowvert
r_I\left(\hat{m}^{ref(c)}_c,\pi^{ref(c)}_c\right) \Arrowvert^2_{\Sigma_I}
\label{eq1}
\end{split}
\end{align}

From the probilistic view \cite{forster2015imu}, a factor graph encodes the posterior probability of the variables, given the available measurements :
\begin{align}
\begin{split}
p(\pi_k|\mathcal{M}_k) &\propto p(\pi_0)p(\mathcal{M}_k|\pi_k)\\ &=p(\pi_0)\prod_{(i,j)\in \mathcal{K}_k} p(\mathcal{M}_I,\mathcal{M}_{IMU}|\pi_k)\\
&=p(\pi_0)\prod_{k\in \mathcal{S}_{IMU}} p(\mathcal{M}_{IMU}|\pi_k)\prod_{c\in \mathcal{C}_{I}} p(\mathcal{M}_I|\pi_k)
\label{eq2}
\end{split}
\end{align}

We assume all the distributions to be Gaussian. The MAP estimate corresponds to the minimum of the negative log-posterior. The negative log-posterior can be written as a sum of squared residual errors:
\begin{align}
\begin{split}
\mathcal{X}_k^{\star}&=\argmin_{\pi_k} ~ -\log_e p(\pi_k|\mathcal{M}_k)\\
&=\argmin_{\pi_k} \Arrowvert r_0\Arrowvert^2_{\Sigma_{0}}+
\sum_{k\in \mathcal{S}_{IMU}}\Arrowvert r_{IMU} \Arrowvert^2_{\Sigma_{IMU}}+
\sum_{c\in \mathcal{C}_I}\Arrowvert r_I \Arrowvert^2_{\Sigma_{I}}
\label{eq3}
\end{split}
\end{align}


Combining the results from the IMU integration (Sect. \ref{IMU}) and the direct image
alignment (Sect. \ref{cam}), and ignoring the prior term, we optimize Eq. \ref{eq1} iteratively using the
Gaussian-Newton method. To solve this type of minimization problem
with a cost function:
\begin{align}
\begin{split}
F(x)=\sum r(x)^2
\label{eq4}
\end{split}
\end{align}

\noindent We introduce the Jacobian and add the weights, which are the inverse of $ \Sigma $, to the equation before we vectorize the cost function as:
\begin{align}
\begin{split}
F(\hat{x}+\Delta x)&=r(\hat{x}+\Delta x)^TWr(\hat{x}+\Delta x)\\
&=(r+J_m\Delta x)^TW(r+J_m\Delta x)\\
&=r^TWr+2r^TWJ_m\Delta x+\Delta x^TJ_m^TWJ_m\Delta x
\label{eq5}
\end{split}
\end{align}
\noindent Then the solution of the minimization is obtained as:
\begin{align}
\begin{split}
J_m^TWJ_m\Delta x&=-J_m^TWr\\
H\Delta x&=-b\\
\Delta x &=-H^{-1}b
\label{eq6}
\end{split}
\end{align}

\subsection{IMU Measurement}
\label{IMU}
Usually, the frequency of the IMU is higher than that of the camera. There are tens of IMU measurement between the two consecutively image interval. The IMU pre-integration between two images provides a prior for image alignment, and also works as a different kind of constraint within the factor graph. The pre-integration $\hat{m}^k_{k+1}$ is given by:
\begin{align}
\begin{split}
\hat{m}^k_{k+1}=\renewcommand{\arraystretch}{1.5}
\begin{bmatrix}
\hat{p}^k_{k+1}\\ \hat{v}^k_{k+1}\\ \hat{\theta}^k_{k+1}\\
\hat{b}^{k+1}_a\\ \hat{b}^{k+1}_g \end{bmatrix}=
\renewcommand{\arraystretch}{1.5}\begin{bmatrix}
\sum^{k+1}_{i=k}\{\frac{1}{2}\hat{R}^k_i(\hat{a}^i_i+b^i_a-R^i_Gg^G)dt^2+\hat{v}^k_idt\}\\
\sum^{k+1}_{i=k}\hat{R}^k_i(\hat{a}^i_i+b^i_a-R^i_Gg^G)dt\\
log(\Pi^{k+1}_{i=k}exp([\hat{\omega}^i_i+b^i_g]_{\times}dt))^{\vee}\\
\hat{b}^k_a+\eta^k_adt\\ \hat{b}^k_g+\eta^k_gdt \end{bmatrix}
\label{eq7}
\end{split}
\end{align}
\noindent where  $g^G$ is the gravity, and
$[\text{ }]_{\times}$ is the operator for skew-symmetric matrix. $\hat{a}^i_i$ and $\hat{\omega}^i_i$ are the IMU
measurements, $\eta^k_a$ and $\eta^k_g$ are white noise affecting the
biases, $\hat{b}^k_a$ and $\hat{b}^k_g$. The biases are initially set
to zeros and the optimized values computed at each subsequent step are
used for the next pre-integration. Through the IMU propagation \cite{ling2016}, we can get the
covariance:
\begin{align}
\begin{split}
\Sigma_{IMU}^{k+1} = F_d (\hat{m}^k_{k+1}) \Sigma_{IMU}^{k} F_d^{T}
(\hat{m}^k_{k+1}) +  G(\hat{m}^k_{k+1})QG^{T}(\hat{m}^k_{k+1})
\label{eq8}
\end{split}
\end{align}

\noindent where $F_d (\hat{m}^k_{k+1})$ is the discrete-time error
state transition matrix, $G(\hat{m}^k_{k+1})$ is the noise transition
matrix, and $Q$ contains all the noise covariance.
\begin{align}
\begin{split}
F_d (\hat{m}^k_{k+1})=
\setlength{\arraycolsep}{3pt}\renewcommand{\arraystretch}{1.5}\begin{bmatrix} I & dt & -\frac{1}{2}\lfloor
R^k_{k+1}(a+\hat{b}^{k+1}_g)\rfloor_{\times}dt^2 & -\frac{1}{2}R^k_{k+1}dt^2 & 0 \\ 
0 & I & -\frac{1}{2}\lfloor
R^k_{k+1}(a+\hat{b}^{k+1}_g)\rfloor_{\times}dt &
-R^k_{k+1}dt & 0 \\ 
0 & 0 & -R^{k+1}_kdt &
0 & -R^k_{k+1}dt \\ 
0 & 0 & 0 & I & 0\\ 
0 & 0 & 0 & 0 & I  \end{bmatrix}
\label{eq9}
\end{split}
\end{align}


Then, the residual function between the IMU pre-integration and the
states is obtained as:
\begin{align}
\begin{split}
r_{IMU}(\hat{m}^k_{k+1},\pi^k_{k+1})=\renewcommand{\arraystretch}{1.5}\begin{bmatrix}
R^k_G(p^G_{k+1}-p^G_k-v^G_kdt)\\ R^k_G(v^G_{k+1}-v^G_k)\\
R^k_GR^G_{k+1}\\ b^{k+1}_a\\ b^{k+1}_g\end{bmatrix}
\ominus \renewcommand{\arraystretch}{1.5}\begin{bmatrix} \hat{p}^k_{k+1}\\ \hat{v}^k_{k+1}\\
\hat{R}^k_{k+1}\\ b^k_a\\ b^k_g \end{bmatrix}
\label{eq10}
\end{split}
\end{align}

We assume that:
\[ R^k_GR^G_{k+1} \ominus \hat{R}^k_{k+1} = log( (\hat{R}^k_{k+1})^TR^k_GR^G_{k+1} )^{\vee}\]
Now the Jacobian of the IMU measurement residual with respect to the error
state is obtained according to the infinitesimal increments in $SO(3)$
\cite{concha2016,xu2016direct} as:
\begin{align}
\begin{split}
J_{IMU}&=
\begin{bmatrix}\dfrac{\partial
	r_{IMU}(\hat{m}^k_{k+1},\pi^k_{k+1})}{\partial
	\delta\pi^G_k}&\dfrac{\partial
	r_{IMU}(\hat{m}^k_{k+1},\pi^k_{k+1})}{\partial \delta
	\pi^G_{k+1}}\end{bmatrix}\\&=
\setlength{\arraycolsep}{0.1pt}\renewcommand{\arraystretch}{1.5}\begin{bmatrix} -R^k_G & -R^k_G & \lfloor
R^k_G(p^G_{k+1}-p^G_k)\rfloor_{\times} & \dfrac{\partial r_{\Delta p^G_k}}{\partial \delta b^k_a} & \dfrac{\partial r_{\Delta p^G_k}}{\partial \delta b^k_g} & R^k_G & 0 & 0 &
0 & 0\\ 0 & -R^k_G & \lfloor R^k_G(v^G_{k+1}-v^G_k)\rfloor_{\times} &
\dfrac{\partial r_{\Delta v^G_k}}{\partial \delta b^k_a} & \dfrac{\partial r_{\Delta v^G_k}}{\partial \delta b^k_g} & 0 & R^k_G & 0 & 0 & 0\\ 0 & 0 & -R^{k+1}_G R^G_k &
\dfrac{\partial r_{\Delta R^G_k}}{\partial \delta b^k_a} & \dfrac{\partial r_{\Delta R^G_k}}{\partial \delta b^k_g} & 0 & 0 & I & 0 & 0\\ 0 & 0 & 0 & -I & 0 & 0 & 0 & 0
& I & 0\\ 0 & 0 & 0 & 0 & -I & 0 & 0 & 0 & 0 & I \end{bmatrix}
\label{eq11}
\end{split}
\end{align}
The details of the calculation will be defined in the appendix \ref{appendix}.

\subsection{Visual Measurement}
\label{cam}
Once the IMU integration is complete, we put the images into two
categories. A {\it key frame} maintains a map (or point clouds) of its environment
and works as a reference to track the subsequent, {\it regular
	frames}. A new image frame is categorized as a key frame when it
overlaps with the current key frame less than a threshold
or the estimated distance between the two frames is over a predefined
value.
With one monocular camera, the inverse depth estimation is adapted for point tracking. the depth map for the new key frame is initialized from the estimation of DSO.

When the system finishes processing a new key frame, subsequent regular frames
are tracked based on the lastest key frame as a reference. We
iteratively minimize the sum of the intensity differences $r_{ij}$ for
all the pixels in the frame to get the relative transformation from
the key frame to the current frame as:
\begin{align}
\begin{split}
\hat{m}^{ref(c)}_c&=\renewcommand{\arraystretch}{1}\begin{bmatrix}\hat{p}^{ref(c)}_c\\ 0\\
\hat{R}^{ref(c)}_c\\ 0\\
0\end{bmatrix}=argmin\sum_i\sum_jr_{ij}\left(\hat{m}^{ref(c)}_c\right)^2
\label{eq12}
\end{split}
\end{align}
\begin{align}
\begin{split}
r_{ij}\left(\hat{m}^{ref(c)}_{c}\right)&=I_{ref(c)}(u_{ij})-I_c(w(R^c_{ref(c)}w^{-1}(u_{ij},d_u)\\
&+p^c_{ref(c)}))
\label{eq13}
\end{split}
\end{align}

\noindent where $I(u_{ij})$ denotes the intensity of the
pixel in position $u_{ij}$ and $d_u$ is the depth of the
pixel. $w(\text{ })$ is the function that project the 3-dimensional
point onto the image plane, while $w^{-1}(\text{ })$ is its inverse projection function. 

After obtaining the optimal visual measurement $\hat{m}^{ref(c)}_c$, we
can compute the residual function based on the current states and the dense tracking result in the reference frame's coordinate:
\begin{align}
\begin{split}
r_I\left(\hat{m}^{ref(c)}_c,\pi^{ref(c)}_c\right)=\renewcommand{\arraystretch}{1.5}\begin{bmatrix}R^{ref(c)}_G(p^G_{ref(c)}-p^G_c)\\0\\R^{ref(c)}_GR^G_c\\0\\0\end{bmatrix}
\ominus\hat{m}^{ref(c)}_c
\label{eq14}
\end{split}
\end{align}

\noindent Then the Jacobian matrix of the dense tracking residual with respect to the 15 states is a sparse matrix:
\begin{align}
\begin{split}
J_I&=\begin{bmatrix}\dfrac{\partial
	r_I\left(\hat{m}^{ref(c)}_c,\pi^{ref(c)}_c\right)}{\partial\delta\pi^G_c}&\dfrac{\partial r_I\left(\hat{m}^{ref(c)}_c,\pi^{ref(c)}_c\right)}{\partial\delta\pi^G_{ref(c)}}\end{bmatrix}\\\stepcounter{equation} 
&=\setlength{\arraycolsep}{0.1pt}\renewcommand{\arraystretch}{1.5}\begin{bmatrix} -R^{ref(c)}_G&0&0&0&0&R^{ref(c)}_G&0&\lfloor
R^{ref(c)}_G(p^G_{ref(c)}-p^G_c)\rfloor_{\times}&0&0\\
0&0&0&0&0&0&0&0&0&0\\ 0&0&I&0&0&0&0&-R^c_GR^G_{ref(c)}&0&0\\
0&0&0&0&0&0&0&0&0&0\\ 0&0&0&0&0&0&0&0&0&0\end{bmatrix}
\label{eq15}
\end{split}
\end{align}

\section{Dense mapping}\label{densemapping}

Rapid camera motions require high-frequency map updates
for the camera to be tracked successfully. According to \cite{engel2016direct}, our system maintains a semi-dense map of high gradient
pixels that is quickly updated and serves for camera
tracking. 

Point candidates are
tracked in subsequent frames using a discrete search along
the epipolar line, minimizing the photometric error.
From the best match, we compute the depth and associated
variance, which is used to constrain the search interval for
the subsequent frame. This tracking strategy is inspired by
LSD-SLAM. Note that the computed depth only serves as
initialization once the point is activated.
\begin{align}
\begin{split}
\hat{d}_u&=argmin\sum_i\sum_jr_{ij}^2
\label{eq16}
\end{split}
\end{align}
\begin{align}
\begin{split}
r_{ij}&=I_{ref(c)}(u_{ij})-I_c(w(R^c_{ref(c)}w^{-1}(u_{ij},d_u)
+p^c_{ref(c)}))
\label{eq17}
\end{split}
\end{align}

\noindent where $I(u_{ij})$ denotes the intensity of the
pixel in position $u_{ij}$ and $d_u$ is the depth of the
pixel. $w(\text{ })$ is the function that project the 3-dimensional
point onto the image plane, while $w^{-1}(\text{ })$ is its inverse projection function. 

When a new keyframe is established, all active points are projected into it and slightly
dilated, creating a semi-dense depth map. 
The selected points with initialied reverse depth are
added to the optimization, we choose a number of candidate
points (from across all keyframes in the fixed window of the \cite{engel2016direct})
and add them into the optimization.

\begin{figure}[h!]
	\begin{center}
		
		\includegraphics[width=0.9\textwidth]{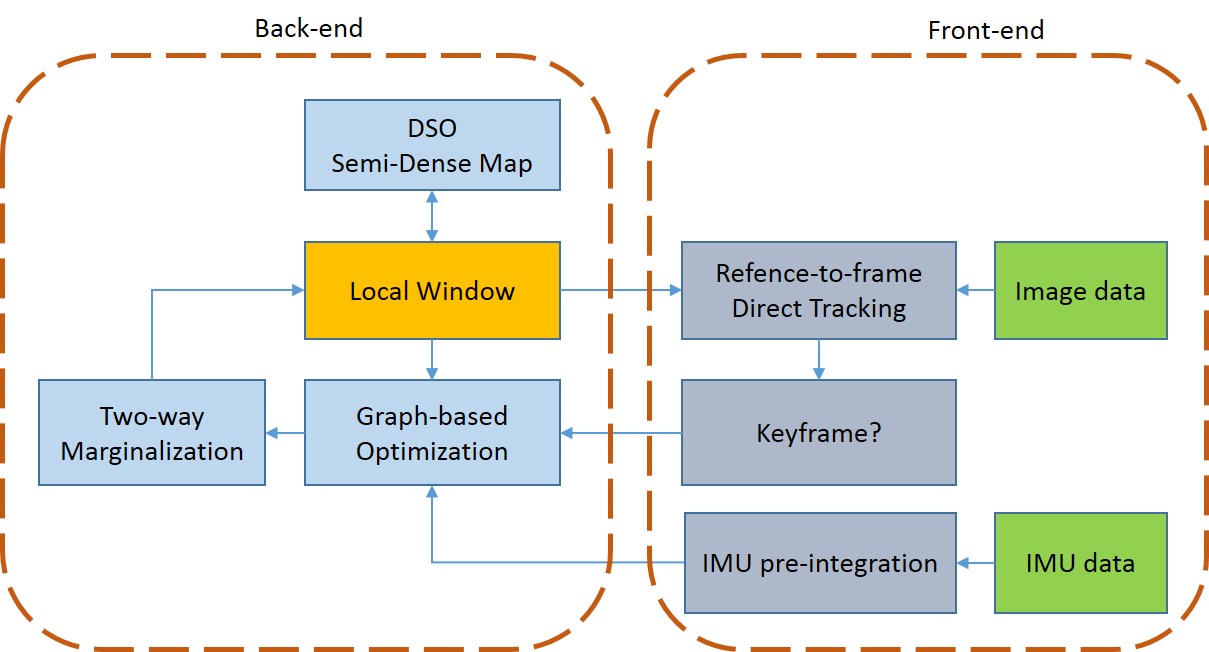}

	\end{center}
	\caption{The pipeline of the proposed system, which comprises of the front-end for direct dense tracking and the back-end for optimization.}
	\label{f3}
\end{figure}

The pipeline of the proposed system is illustrated in Figure \ref{f3}. Between the time interval of two consecutive images, IMU data from the sensor is first pre-integrated. In the front-end, the dense tracking
thread obtains incremental camera motion using a direct
keyframe-to-frame dense tracking algorithm, assisted by IMU pre-integration. Based on the instant tracking performance, it also determines whether to add the frame as a keyframe or regular frame, or report tracking failure. If a keyframe is added, a depth map will be generated from the DSO. The back-end periodically checks the frame list buffer. The new frame and IMU measurements are added to the optimization thread.
If a keyframe is added, Graph optimization
is then applied to find the maximum a posteriori
estimate of all the states within the sliding window using
connections from IMU pre-integration, dense tracking and the prior. A two-way marginalization scheme
that selectively removes states is performed in order to
both bound the computational complexity and maximize the
information stored within the sliding window.

\section{Robust method} \label{implement}

\subsection{Robust Norm}
Image processing suffers outliers commonly because of image noise and ambiguity. $L_2$ norm is sensitive to outliers. Very small number of outliers will drive the solution away.
Other researchers choose the Huber function \cite{engel2013semi}. It gives small residuals quadratic
and large residuals only linear influence. In contrast to the Tukey function, it is convex and
therefore does not introduce new local minima to the optimization. The Huber function is
defined as:
\begin{align}
\begin{split}
	\mathrm{\omega_{Huber}(x)} = 
	\begin{cases}
	1  ~~~~~~~~~~~if~ |x|\leq0;\\
	\dfrac{k}{|x|}~~~~~~~~ otherwise.\\
    \end{cases}
\label{eq18}
\end{split}
\end{align}

\noindent where $x$ is the error and $k$ is a given threshold.
\subsection{Keyframe and Point Management}
The front-end of visual odomerty needs to determine the active frame / point set, provide initialization for new parameters and decide when a frame / point should be marginalized  \cite{engel2016direct} so as to make the system computationally efficient and accurate. Our keyframe and point management is largely based on the DSO \cite{engel2016direct} except for that we add two more continuous regular frames into the sliding window \cite{shen2015tightly}. This two regular frames are connected by the IMU pre-integration.

\subsection{Two-way Marginalization}
During the graph optimization, the states set increases with any new frame added so as to require more memory and computational resources. In order to reduce the computational complexity, we need to marginalize states based on a two-way marginalization scheme \cite{shen2015tightly,ling2016,OKVIS} to maintain a sliding window of states and convert measurements
corresponding to the marginalized states into a prior.
By two-way marginalization, all information of
the removed states is kept and the computation complexity is
bounded, which is fundamentally different from the traditional
keyframe-based approaches that simply drop non-keyframes.
Front marginalization removes the second newest frame, ensuring that the time interval for each IMU preintegration is bounded in order to bound the accumulated error. While back marginalization removes the oldest keyframes
within the sliding window to improve the effectiveness of multi-constrained factor graph optimization since the dense alignment and the IMU pre-integration depend on
whether an older state is kept within the sliding window. In this situation, the oldest keyframes provide very few information.

Intuitively, the two-way marginalization will remove the second newest state if the frame tracking is reliable. The second newest state will be marginalized in the next
round if the dense tracking is good and the second newest
state is near to the current keyframe. Otherwise, the oldest
state will be marginalized. The distance is thresholded by
a weighted combination of translation and rotation between
the latest keyframe and the second newest frame.

For the prior matrix $ \{\Lambda_p,b_p\}$, which contains the information from the marginalized states:
\begin{align}
\begin{split}
\Lambda_p &= \Lambda_p + \sum_{k\in \mathcal{S}^-_{IMU}}(J_{k+1}^k)^T(\Sigma_{k+1}^k)^{-1}J_{k+1}^k\\
&+ \sum_{c\in \mathcal{C}^-_I}(J_I^{ref(c)})^T(\Sigma_I^{ref(c)})^{-1}J_I^{ref(c)}
\label{eq19}
\end{split}
\end{align}
where $\mathcal{S}^-_{IMU}$ and $\mathcal{C}^-_I$ are the set of removed IMU and visual measurement, respectively. The prior is then marginalized via the Schur complement \cite{sibley2010}.
\begin{figure}[t!]
	\begin{center}
		\includegraphics[width=1.\textwidth]{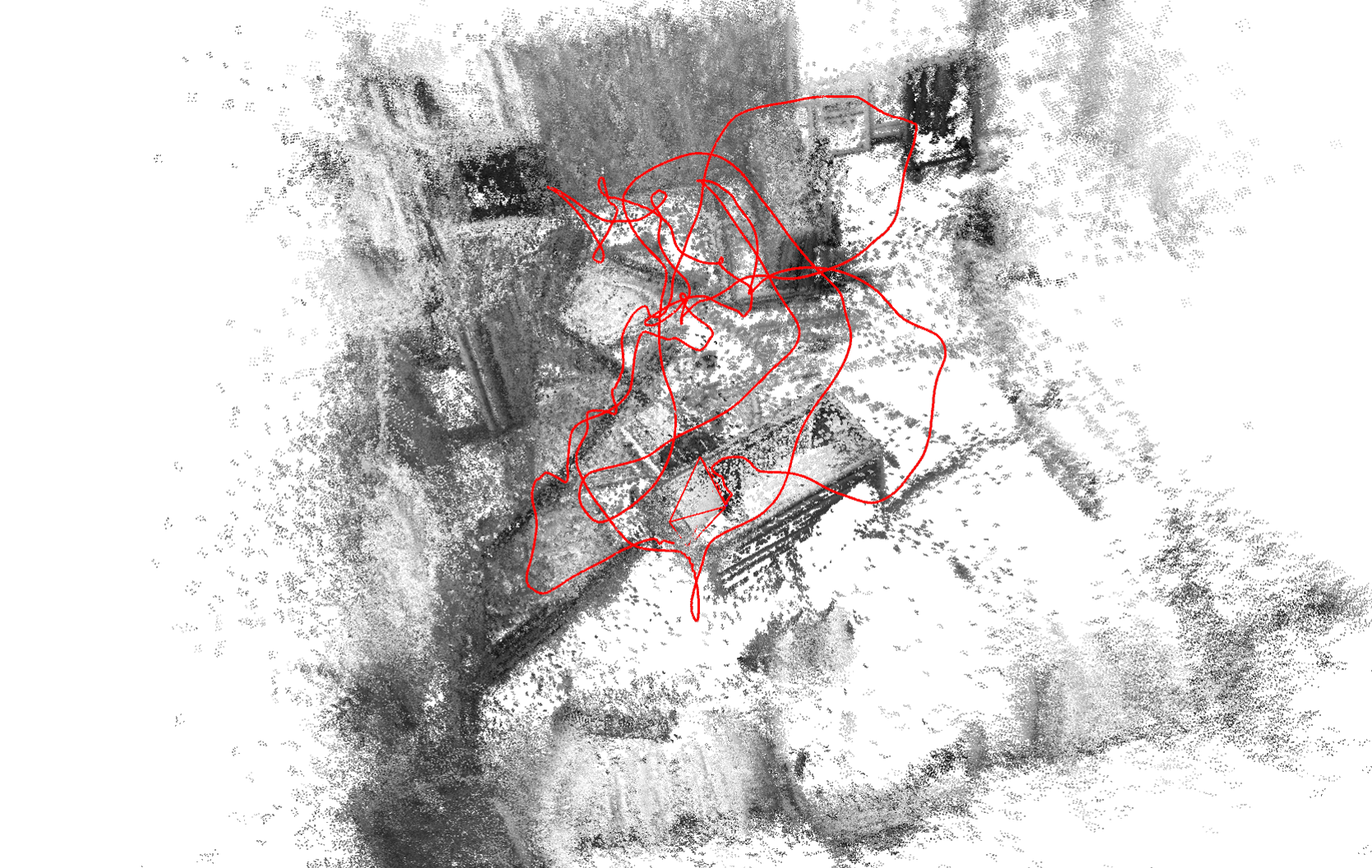}

	\end{center}
	\caption{3D reconstruction on $V1\_01$ sequence. The red line is the estimated trajectory while the black part is the 3D point cloud.}
	\label{f4}
\end{figure}
\begin{figure}[h!]
	\begin{center}
		
		\includegraphics[width=1.0\textwidth]{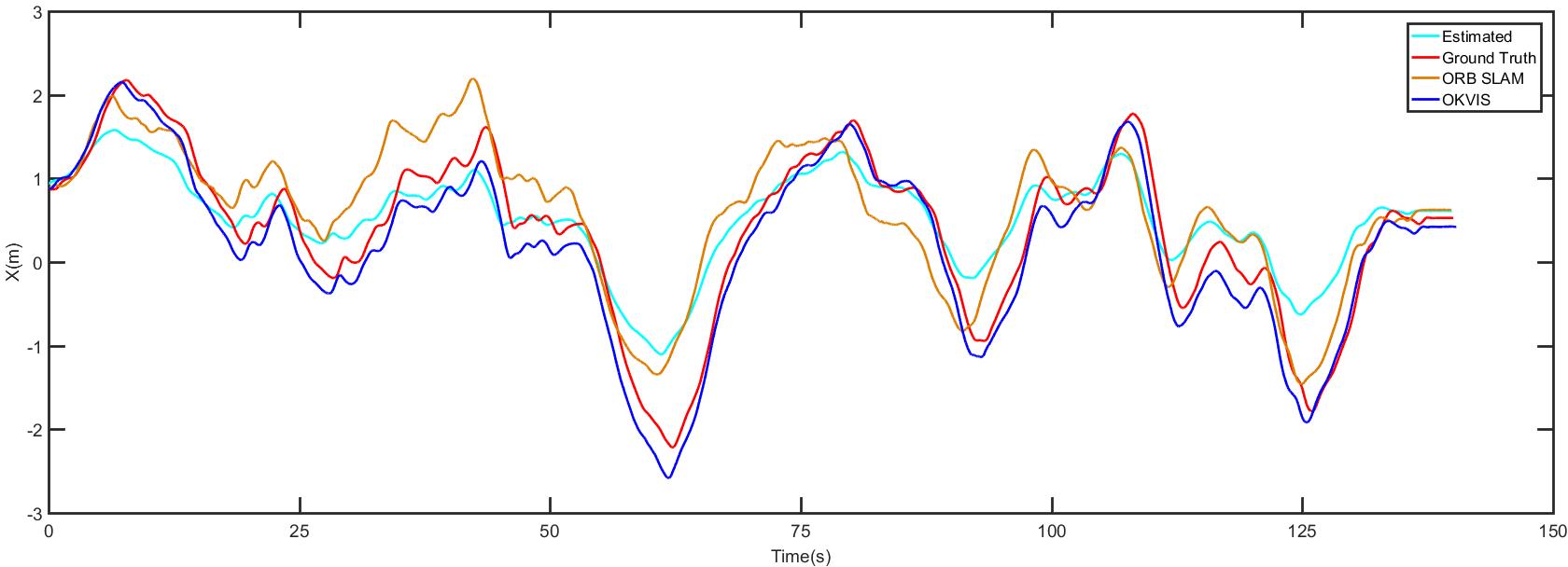}
		
		\includegraphics[width=1.0\textwidth]{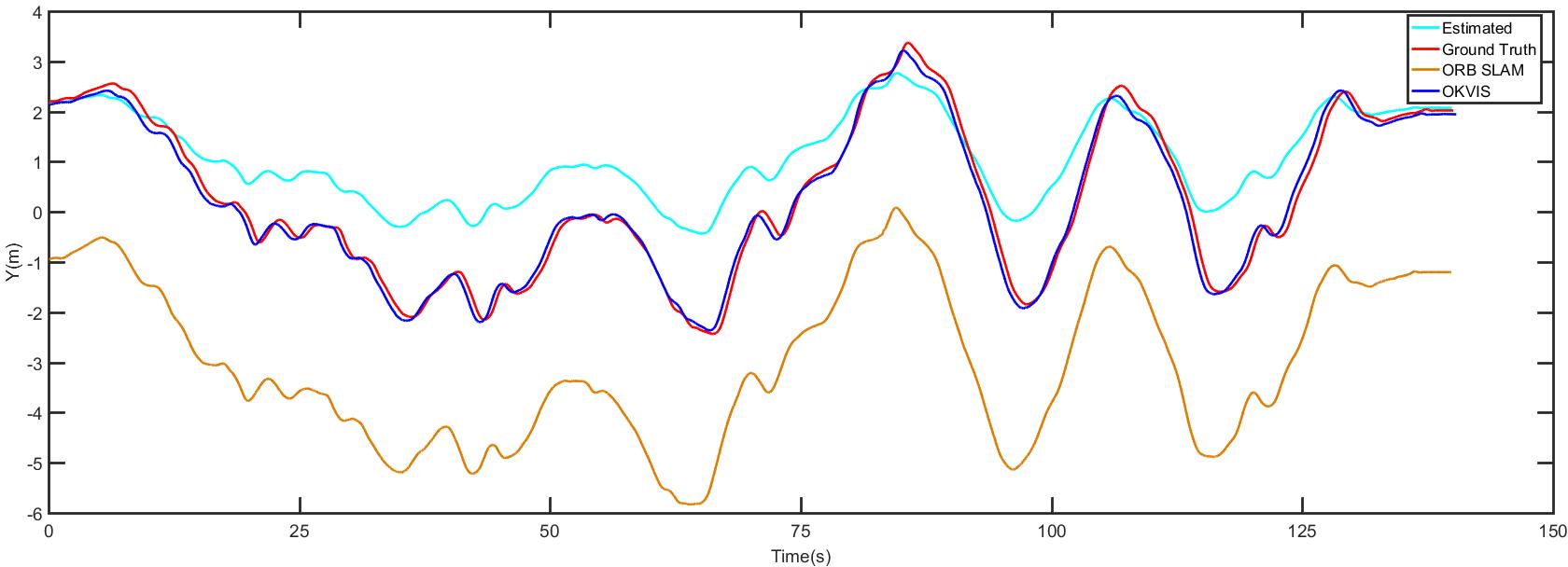}
		
		\includegraphics[width=1.0\textwidth]{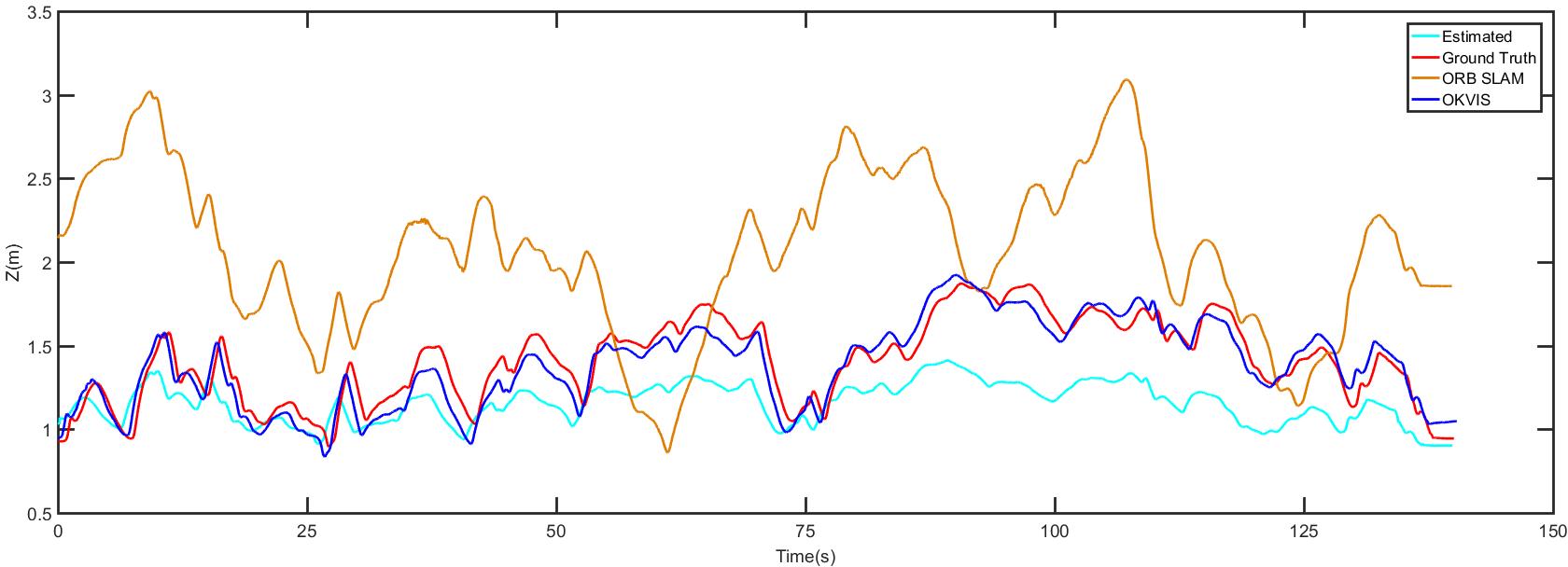}
	\end{center}
	\caption{Comparison of trajectory estimations between our
		algorithm, the ground truth for
		the sequence $V1\_01$, OKVIS, and ORB SLAM.}
	\label{f5}
\end{figure}
\begin{figure}[t!]
	\begin{center}
		
    \hspace*{-1cm}\includegraphics[width=1.2\textwidth]{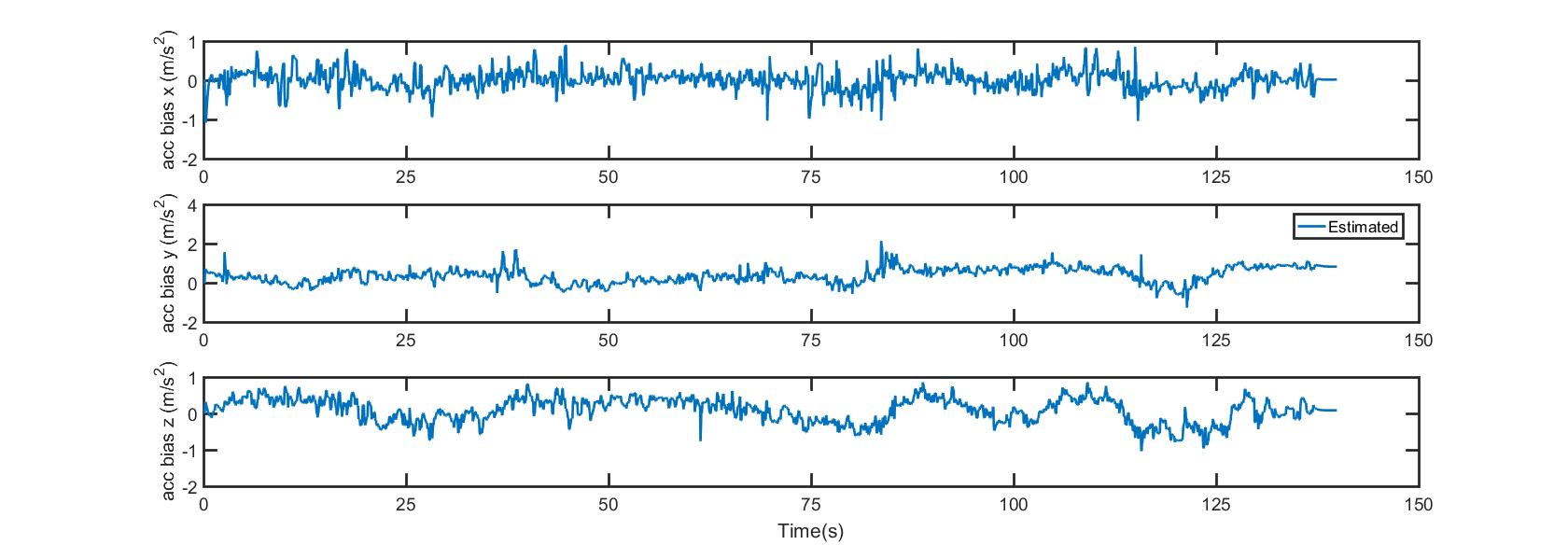}
	\hspace*{-1cm}\includegraphics[width=1.2\textwidth]{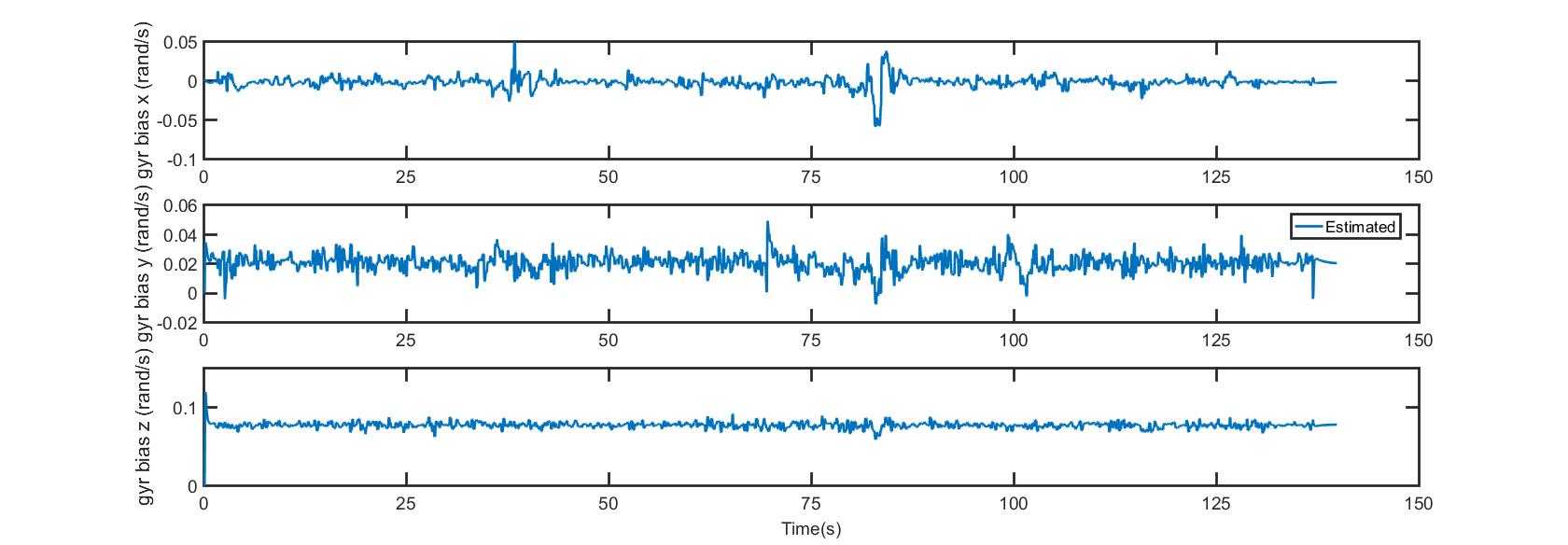}
		
	\end{center}
	\caption{IMU acceleration bias and gyroscope bias estimation results from our algorithm.}
	\label{f6}
\end{figure}
\section{Experiment}\label{experiment}
In the literature, a bunch of motion tracking algorithms has been suggested; It could be visual only odometry, loosely coupled fusion, or tightly coupled fusion. How they perform in relation to each
other, however, is often unclear, since results are typically shown on individual datasets with different motion and lighting characteristics. In order to make a strong argument for our presented work, we will thus compare it to state-of-the art odometry methods.

\subsection{Experiment Setup}
We evaluated our system on two different sequences of the publically available popular dataset. The dataset is the the European
Robotics Challenge (EuRoC) dataset \cite{EUROC}, which contains 11 visual-inertial sequences recorded in 3 different indoor environment. The datasets are increasingly difficult to process in terms of flight dynamics and lighting conditions. The two sequences that we use are the V1$\_$01 and the MH$\_$02, which could be successfully handled and compared with other methods, and they are representative sequences of the dataset. The entire algorithm is developed in C++ using ROS. 

\subsection{Evaluation on EuRoC dataset}
\begin{figure}[h!]
	\begin{center}
		
		\includegraphics[width=1.0\textwidth]{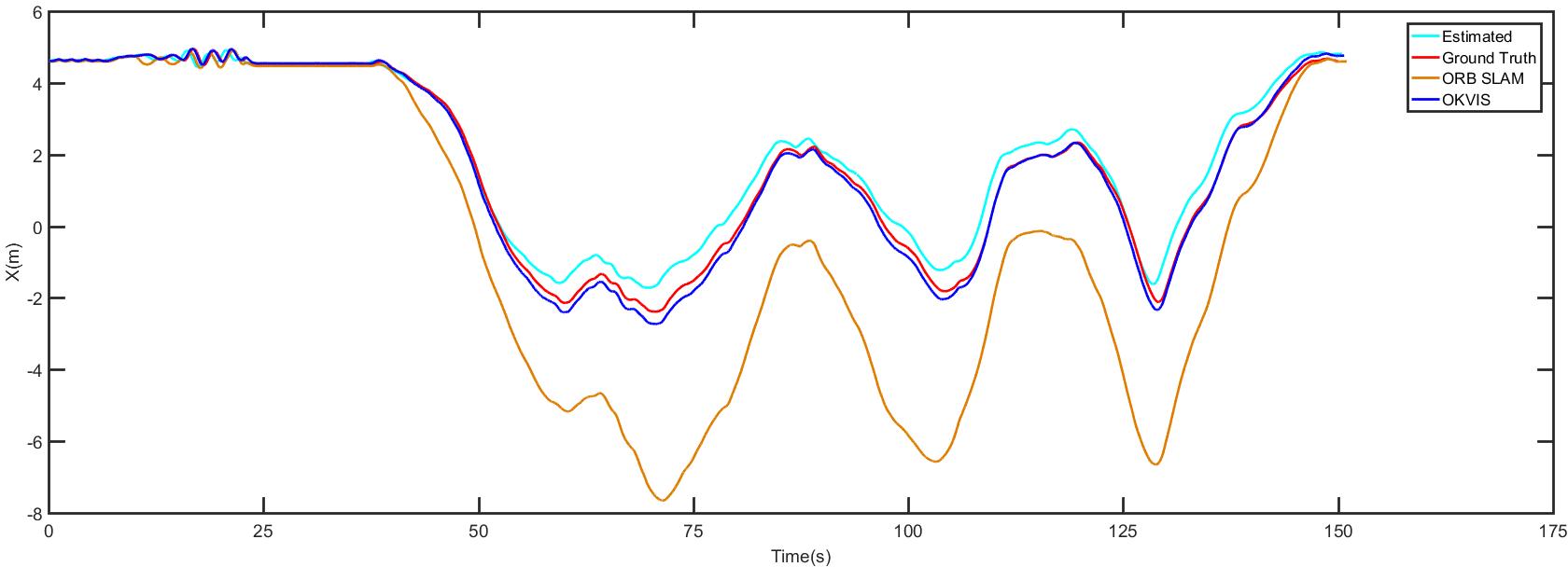}
		
		\includegraphics[width=1.0\textwidth]{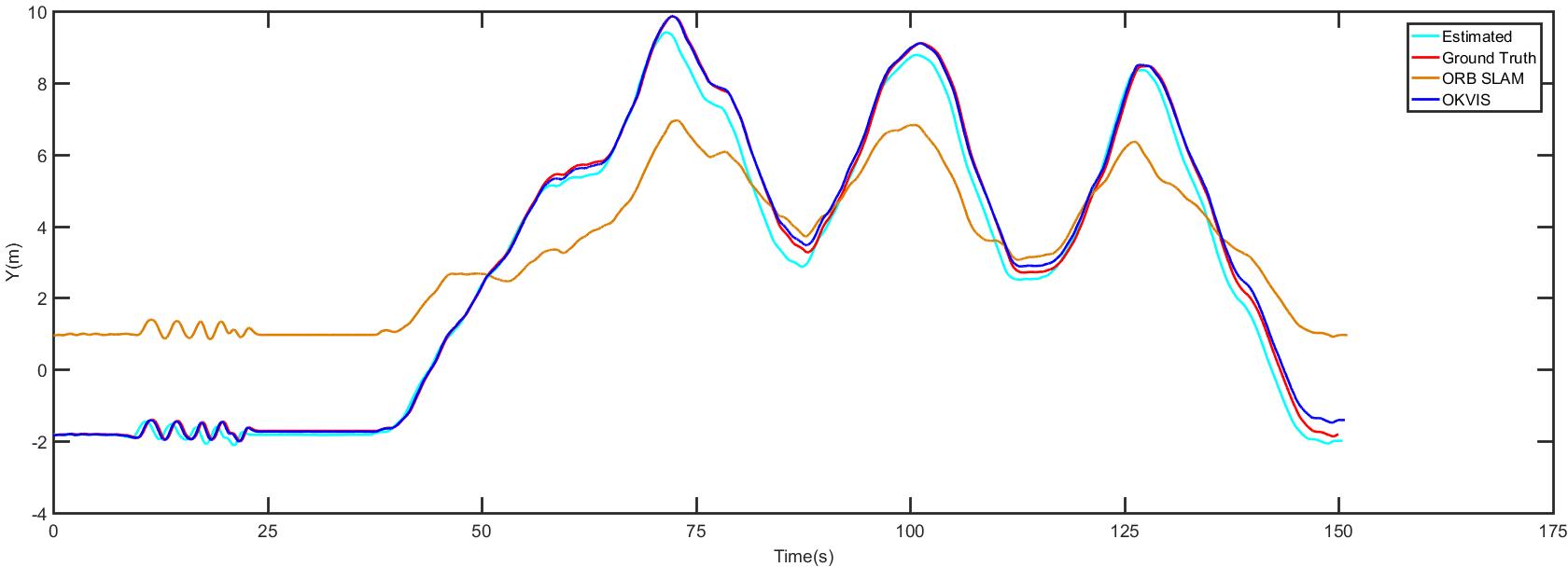}
		
		\includegraphics[width=1.0\textwidth]{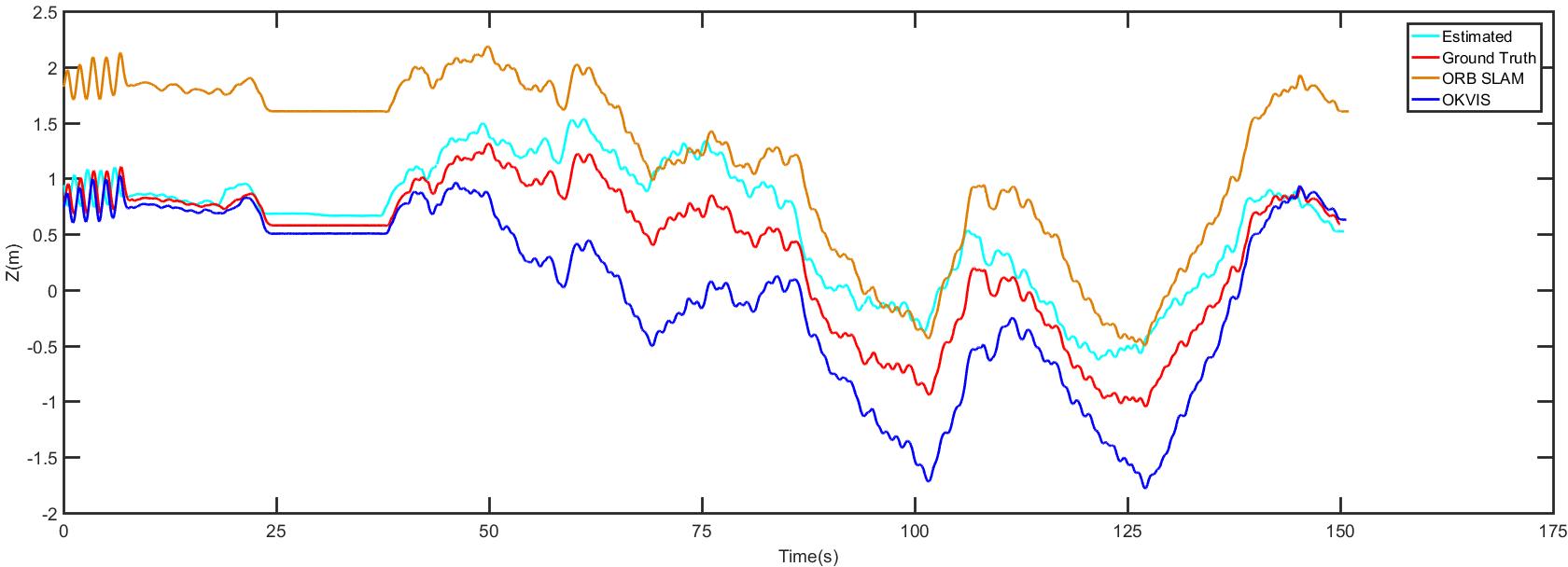}
	\end{center}
	\caption{Comparison of trajectory estimations between our
		algorithm, the ground truth for
		the sequence $MH\_02$, OKVIS, and ORB SLAM.}
	\label{f8}
\end{figure}

The European provides the ground
truth data at $1mm$ accuracy. We compared the performance of our
system to the ground truth. Since the sequences are
recorded in different coordinate systems, we transform the results
to the coordinate system of the ground truth. Figure~\ref{f8} adn Figure~\ref{f5}
shows the comparison for position.

In addition, the 3-dimensional reconstruction of the surroundings
obtained from monocular configurations for this
dataset are shown in
Figure~\ref{f1}. It is evident that the reconstruction from the
stereo setup is smoother and the contour of the structure in the map
is clearer.

\begin{figure}[h!]
	\begin{center}
		
		\includegraphics[width=0.9\textwidth]{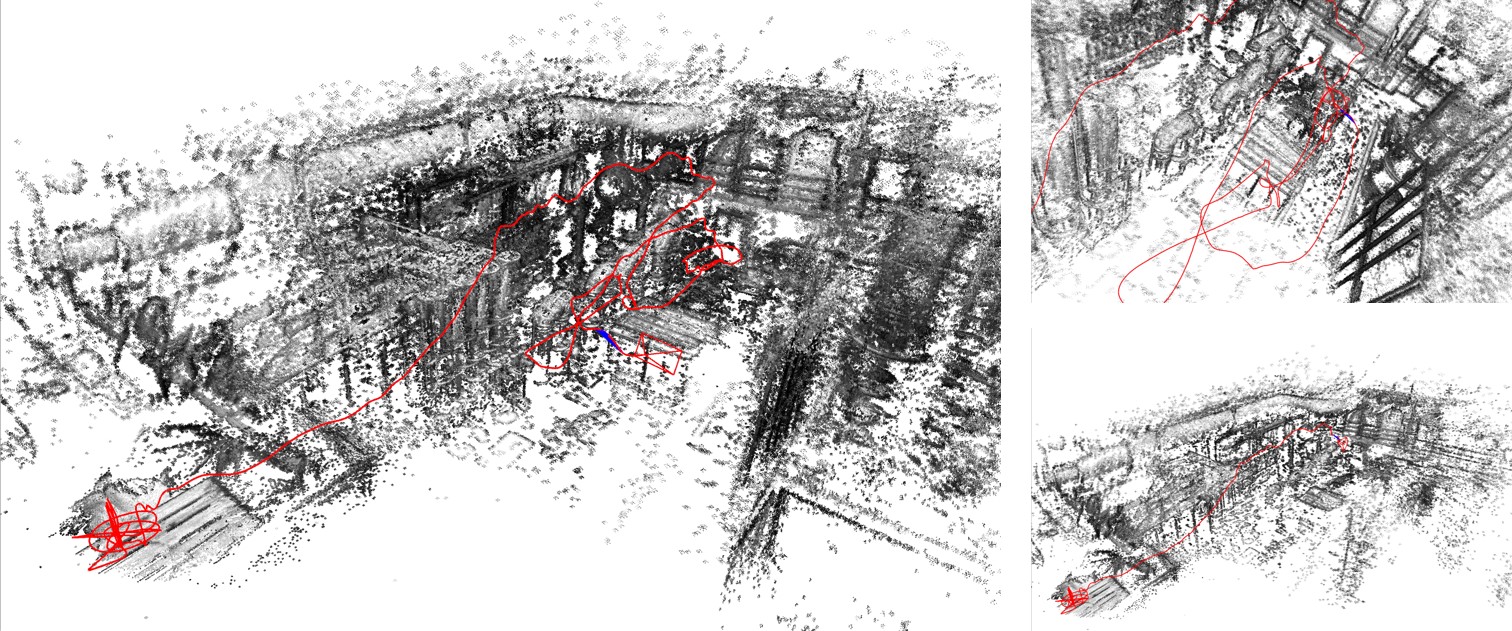}
	\end{center}
	\caption{Whole view of 3D reconstruction on $MH\_02$ sequence. The red line is the estimated trajectory while the black part is the 3D point cloud. The left part shows the reconstruction of the machine house. The right two images show the details from different view angular.}
	\label{f7}
\end{figure}

To evaluate the efficiency of the proposed system, two different methods dealing with the odometry problem are used for comparison. The OKVIS\cite{OKVIS} implements the visual-inertial odometry in a tightly coupled fusion method that achieves the state-of-the-art performance, and ORB SLAM \cite{mur2015orb} uses the ORB features for tracking and mapping.

Figure {\ref{f5}} shows the comparison of trajectory estimations tested on the $V1\_01$ sequence. The dataset is characterized by fast motion and image blur. The ORB SLAM runs for visual odometry without IMU measurements to provide up-to-scale information. This results into scale drift for three directions. While OKVIS preforms the best to be the closest to the ground truth since it fuses the IMU measurements with the dense tracking results and takes the depth of features into the optimization formulation. Figure {\ref{f6}} is the estimated biases for the IMU. In our method, We perform a semi-dense reconstruction and do not take the estimated feature depth into the optimization formulation. And due to the depth estimation error from the DSO, the visual inertial fusion for motion tracking could lead to a reduced performance. Nevertheless, ORB SLAM fails to track in the two sequences with up-to-scale accuracy. The other approaches are able to track these sequences successfully. The similar trajectory estimation results on $MH\_02$ sequence are showed in Figure {\ref{f8}}.

For the semi-dense mapping, Figure {\ref{f7}} is the reconstruction of a machine house from the $MH\_02$ which is an indoor dark scene. Figure {\ref{f4}} demonstrates the reconstruction of the environment in a small room.

As demonstrated by the challenging experiments, the proposed system is robust and able to handle fast motion and different lighting condition, but it is still subject to scale drift. In all case, the inertial measurement of the IMU provides constraints to largely eliminate the drifts. 

\section{Conclusion} \label{conclusion}
We propose a direct visual-inertial odometry and mapping that tracks the camera
motion and reconstructs the environment into a 3-dimensional model. Our
method is based on a novel integration of dense image tracking and
optimization of tracking error with the IMU measurements. Experimental
results on the popular dataset demonstrated the performance of our
method and the potential for application in unmanned vehicle.
\clearpage
\bibliographystyle{splncs04}
\bibliography{mybibfile}

\clearpage
\appendix
\section*{Appendix}
\label{appendix}
\section{Jacobians of the IMU residual with respect to the IMU parameters}
\subsection{Background}\label{sec}
We take the preintegrated measurements $\Delta p^k_G$ $\Delta v^k_G$ $\Delta R^k_G$ as:
\begin{align*}
\Delta p^k_G&=\frac{1}{2}\sum^{k+1}_{i=k}\{2(N-i)+1\}R^k_i(\hat{a}^i_i+b^i_a-R^i_Gg^G)dt^2 +Nv_k^Gdt\\
\Delta v^k_G&=\sum^{k+1}_{i=k}R^k_i(\hat{a}^i_i+b^i_a-R^i_Gg^G)dt   \\
\Delta R^k_G&=\prod^{k+1}_{i=k}exp(\hat{\omega}^i_i+b^i_g)dt
\end{align*}

\noindent According to infinitesimal increment in so(3) with right hand-multiplication \cite{PREIN,concha2016}
\begin{equation}
\label{eq22}
exp([\theta +\delta\theta]_{\times}) =  exp([\theta]_{\times})exp([J_r(\theta)^{-1}\delta\theta]_{\times})
\end{equation}
Then, we take the same fist-order approximation for the logarithm introduced in \cite{PREIN}
\begin{equation}
\label{eq23}
log(exp([\theta]_{\times}exp([\delta\theta]_{\times})))^{\vee}=\theta+ J_r(\theta)^{-1}\delta\theta  
\end{equation}
\noindent where $J_r(\text{ })$ is the $SO(3)$ Jacobian.\\
Another efficient relation for linearization is directly from the adjoint representation
\begin{equation}
\label{eq24}
exp([\theta]_{\times})R = R exp([R^T\theta]_{\times})  
\end{equation}

\subsection{Jacobians}
To calculate the Jacobian of the rotation error $r_{\Delta R^G_k}$ with respect to the angular velocity bias $b^i_g$, we apply the three equations in \ref{sec} to move all the increment terms to the right side
\begin{align*}
r_{\Delta R^G_k}(b^i_g+\delta b^i_g) &=log( (\hat{R}^k_{k+1})^TR^k_GR^G_{k+1} )^{\vee}\\ 
&\overset{}{=}log((\prod^{k+1}_{i=k}exp(\hat{\omega}^i_i+b^i_g +\delta b^i_g)dt)^T(R^G_k)^TR^G_{k+1} )^{\vee} 
\end{align*}

\noindent Similarly to the other Jacobians, we apply the equations in \ref{sec} to move all the increment terms to the right side and obtain:
\begin{align*}
\dfrac{\partial r_{\Delta p^k_G}}{\partial \delta b^k_a}&=-\frac{1}{2}\sum^{k+1}_{i=k}\{2(N-i)+1\}R^k_idt^2\\
\dfrac{\partial r_{\Delta v^k_G}}{\partial \delta b^k_a}&=-\sum^{k+1}_{i=k}R^k_idt\\
\dfrac{\partial r_{\Delta R^k_G}}{\partial \delta b^k_a}&=0\\
\dfrac{\partial r_{\Delta p^k_G}}{\partial \delta b^k_g}&=\frac{1}{2}\sum^{k+1}_{i=k}\{C[\hat{a}^i_i+b^i_a]_{\times}Bdt^2\}\\
\dfrac{\partial r_{\Delta v^k_G}}{\partial \delta b^k_g}&=\sum^{k+1}_{i=k}D[\hat{a}^i_i+b^i_a]_{\times}Bdt \\
\dfrac{\partial r_{\Delta R^k_G}}{\partial \delta b^k_g}&=-J_r(r_{\Delta R^k_G})^{-1}exp([r_{\Delta R^k_G}]_{\times})^T\sum^{k+1}_{i=k}\{[\prod^{k+1}_{m=i+1}exp\left([\hat{\omega}^i_i+b^i_g]_{\times}dt\right)]^TJ_r\left((\hat{\omega}^i_i+b^i_g)dt\right)dt\}\\
C&=[\prod^{i-1}_{m=k}\{2(N-i)+1\}exp\left([\hat{\omega}^m_m+b^m_g]_{\times}dt\right)]\\
B&=\sum^{i-1}_{l=k}\{[\prod^{i-1}_{m=l+1}exp\left([\hat{\omega}^m_m+b^m_g]_{\times}dt\right)]^TJ_r\left((\hat{\omega}^l_l+b^l_g)dt\right)dt\}\\
D&=\prod^{k+1}_{m=i+1}exp\left([\hat{\omega}^m_m+b^m_g]_{\times}dt\right)
\end{align*}

\end{document}